\documentclass{ieeetj}
\usepackage{cite}
\usepackage{amsmath,amssymb,amsfonts}
\usepackage{algorithmic}
\usepackage{graphicx,color}
\usepackage{textcomp}
\usepackage{xcolor}
\usepackage{hyperref}
\hypersetup{hidelinks=true}
\usepackage{algorithm,algorithmic}
\def\BibTeX{{\rm B\kern-.05em{\sc i\kern-.025em b}\kern-.08em
    T\kern-.1667em\lower.7ex\hbox{E}\kern-.125emX}}
\AtBeginDocument{\definecolor{tmlcncolor}{cmyk}{0.93,0.59,0.15,0.02}\definecolor{NavyBlue}{RGB}{0,86,125}}

\definecolor{LightRed}{HTML}{FFE0E0}
\newcommand{\ours}{BGM2Pose}
\newcommand{\etal}{\textit{et al.}}
\newcommand{\ie}{\textit{i.e.}}
\def\red#1{\textcolor{Red}{#1}}
\usepackage{multirow}
\usepackage{multicol}
\usepackage[dvipsnames]{xcolor}
\usepackage[table]{xcolor}

\usepackage{booktabs}
\usepackage{type1cm}
\usepackage{comment}

\def\OJlogo{\vspace{-4pt}$<$Society logo(s) and publication title will appear here.$>$}
\def\seclogo{\vspace{10pt}$<$Society logo(s) and publication title will appear here.$>$}

\def\authorrefmark#1{\ensuremath{^{\textbf{#1}}}}

\begin{document}
\receiveddate{XX Month, XXXX}
\reviseddate{XX Month, XXXX}
\accepteddate{XX Month, XXXX}
\publisheddate{XX Month, XXXX}
\currentdate{XX Month, XXXX}
\doiinfo{XXXX.2022.1234567}

\markboth{}{Author {et al.}}

\newcommand{\revm}[1]{{\color{black}#1}}

\title{BGM2Pose: Active 3D Human Pose Estimation with Non-Stationary Sounds}


\author{
  Yuto Shibata\authorrefmark{1},
  Yusuke Oumi\authorrefmark{1},
  Go Irie\authorrefmark{1,2}, \\
  Akisato Kimura\authorrefmark{3},
  Yoshimitsu Aoki\authorrefmark{1},
  and Mariko Isogawa\authorrefmark{1,4}
  }
\affil{Keio University, Yokohama 223-8522, Japan}
\affil{Tokyo University of Science, Katsushika 125-8585, Japan}
\affil{NTT, Inc., Keihanna Science City, Kyoto, 619-0237, Japan}
\affil{JST Presto, Chiyoda 100-0076, Japan}
\corresp{Corresponding author: Yuto Shibata (email: 
yuto071508@keio.jp).}
\authornote{This work was partially supported by JST Presto JPMJPR22C1, JSPS Grant-in-Aid for Challenging Research (Exploratory) 24K22296, JST BOOST, Japan Grant Number JPMJBS2409, and Keio Gijuku Academic Development Funds. 
This paper has supplementary downloadable material available at http://ieeexplore.ieee.org., provided by the author. The material includes a demonstration video showing the results of the proposed 3D human pose estimation method and background music. This material is 108.5 MB in size.}

\begin{abstract}
We propose \ours, a non-invasive 3D human pose estimation method using 
\revm{pre-selected everyday music (e.g., background music)} as active sensing signals. 
Unlike existing approaches that significantly limit practicality by employing intrusive chirp signals within the audible range, our method utilizes \revm{user-selected} natural music that causes minimal discomfort to listeners.
Estimating human poses from standard music presents significant challenges.
In contrast to sound sources specifically designed for measurement, regular music varies in both volume and pitch. These dynamic changes in signals caused by music are inevitably mixed with alterations in the sound field resulting from human motion, making it hard to extract reliable cues for pose estimation.
To address these challenges, \ours~ introduces a Contrastive Pose Extraction Module that employs contrastive learning and hard negative sampling to eliminate musical components from the recorded data, isolating the pose information. Additionally, we propose a Frequency-wise Attention Module that enables the model to focus on subtle acoustic variations attributable to human movement by dynamically computing attention across frequency bands.
Experiments suggest that our method outperforms the existing methods, demonstrating substantial potential for real-world applications. 
Project page: \url{https://yutoshibata07.github.io/bgm2pose-project-page/}
\end{abstract}

\begin{IEEEkeywords}
Human Pose Estimation, Sound, Deep Learning

\end{IEEEkeywords}


\maketitle

\section{INTRODUCTION}
This paper
tackles the challenging task of estimating human poses using common background music played through off-the-shelf speakers and a microphone, without relying on specialized sensing signals.
Human pose estimation and activity recognition are fundamental capabilities with diverse applications, including health monitoring~\cite{
gao2020wearable}, sports performance analysis~\cite{Yeung_2024_CVPR
}, rehabilitation~\cite{cotton2023markerless
}, and XR applications~\cite{tome2019xr}. 
Although many effective methods have been proposed, most of them rely on RGB images or videos captured by standard cameras~\cite{openpose_tpami,stridedTransformer
}. These approaches face challenges such as vulnerability to occlusion and low-light conditions~\cite{lee2023human}, as well as privacy concerns due to the capture of identifiable features like faces~\cite{fredrikson2015model, zheng2023deep}.

\begin{figure*}[t]
\centering

\centerline{\includegraphics[width=1.0\linewidth]{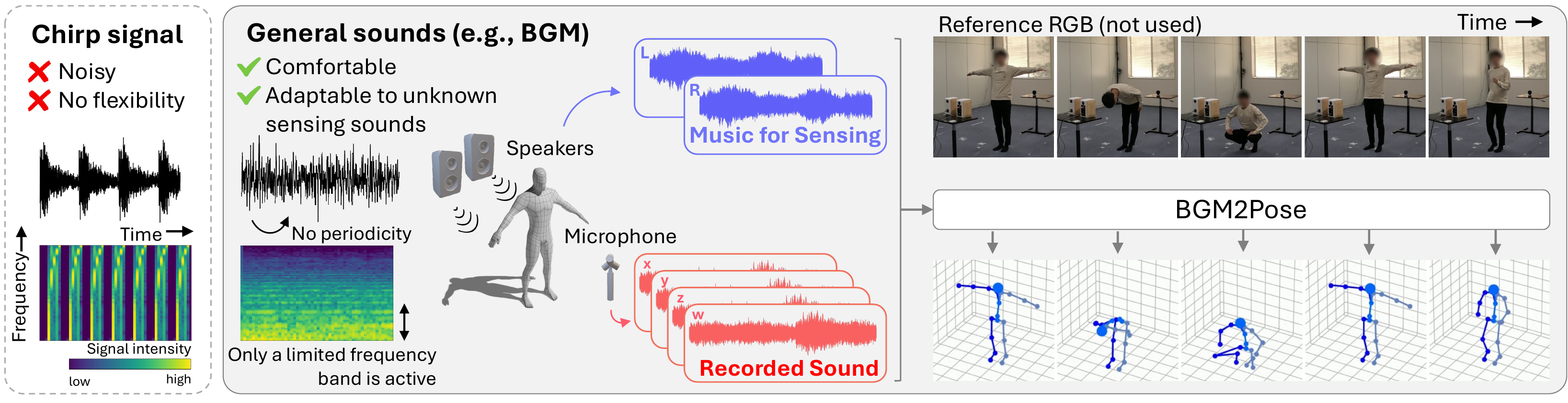}}
\caption{We propose BGM2Pose, a 3D human pose estimation that utilizes background music that naturally integrates into everyday life, for active acoustic sensing. Given playback music and recorded acoustic signals, our method estimates frame-by-frame 3D human poses. Unlike chirp signals, BGM tracks contain unstable sounds and limited frequency bands.
}
\label{fig:teaser}
\vspace{-3mm}
\end{figure*}

Acoustic signals are one promising modality to address these challenges. They are less affected by lighting conditions and, due to their longer wavelengths, are less likely to contain visual privacy-related information such as facial expressions. While other modalities with similar advantages (e.g., millimeter-wave or WiFi signals) also exist~\cite{
ren2022gopose, zhao2018through, jiang2020towards, fan2025diffusion}, these signals often face restrictions in environments with medical equipment or on airplanes. By contrast, sound can be used almost anywhere.

\revm{
Several methods using acoustic signals for human state estimation have been explored. Existing approaches typically employ chirp-based sensing signals, such as Time-Stretched Pulse (TSP) signals, and estimate human states by repeatedly emitting the same signal and analyzing changes caused by human motion. While some acoustic sensing systems~\cite{posekernellifter,mahmud2023posesonic} use ultrasonic chirps (above 18--20\,kHz) that are inaudible to humans, the acoustic pose estimation method most closely related to our task~\cite{shibata2023listening} operates over a wide range of audible frequencies (approximately 20\,Hz to 20\,kHz), making them uncomfortable for prolonged everyday use. Furthermore, chirp-based sensing methods generally require the same sensing signal during inference as was used during training, which restricts their flexibility and practical deployment. These limitations motivate our use of everyday music as the sensing source, enabling more user-friendly and practical acoustic pose estimation.
}

\begin{table*}[t]
    \centering
        \caption{Comparisons between existing human-pose–related studies and our method.}
        \scalebox{0.9}{
        \begin{tabular}{crcccccc}
        \toprule
        Task & Method & Modality & Comfort & Restricted Usage & Inference Signal\\
        \midrule
        \multirow{5}{*}{\shortstack[c]{Pose \\ Estimation}} & RGB-based~\cite{openpose_tpami,stridedTransformer
        } & RGB & \color{Red}{Low} (privacy \revm{concerns}) & \color{Red}{Dark \revm{environment}, occlusion} & - \\ 
        & RF/WiFi-based~\cite{zhao2018through, jiang2020towards, ren2022gopose, Li_2022_WACV} & RF/WiFi & \color{ForestGreen}{High} & \color{Red}{Occlusion, electronic device} & \color{Red}{Fixed}\\
        & Audio2Hand~\cite{audiotouch, 
        handgesture} & Audio & \color{Red}{Low (invasive)} & Soundproof environment & \color{Red}{Fixed}\\
        & Sound2Pose~\cite{posekernellifter,shibata2023listening} & Audio & \color{Red}{Low} (chirp signals) & Soundproof environment & \color{Red}{Fixed}\\
        \rowcolor[rgb]{0.9, 0.9, 0.9}
        & \ours (Ours) & Audio & \color{ForestGreen}{High} & Soundproof environment & \color{Green}{Adaptive}\\
        \midrule
        \multirow{2}{*}{\shortstack[c]{Motion \\ Generation}} & Music2Motion~\cite{tseng2023edge, shlizerman2018audio} & Audio & \color{ForestGreen}{High} & \color{Red}{Inapplicable for sensing} & - \\
        & Speech2Pose~\cite{
        ginosar2019gestures, yi2023generating} & Audio & \color{ForestGreen}{High}& \color{Red}{Inapplicable for sensing} & -\\
        \bottomrule
        \end{tabular}%
        }
    \label{tab:comparison}
    \vspace{-4mm}
\end{table*}

Therefore, this paper addresses these challenges by proposing a task of \textbf{3D Human Pose Estimation Based on Non-Stationary Sounds} (Fig.~\ref{fig:teaser}), which significantly expands the framework of non-invasive estimation of dynamic human pose. In this task, we utilize standard \revm{pre-selected} background music (BGM) as the sensing signal.
With BGM, our system significantly enhances comfort and practicality compared to a noisy, chirp-based framework. It also retains the advantages of acoustic sensing, including robustness to occlusion and dark environments, as well as visual privacy protection.

\begin{figure}[t]
\centering
\vspace{-3mm}
\centerline{\includegraphics[width=0.9\linewidth]{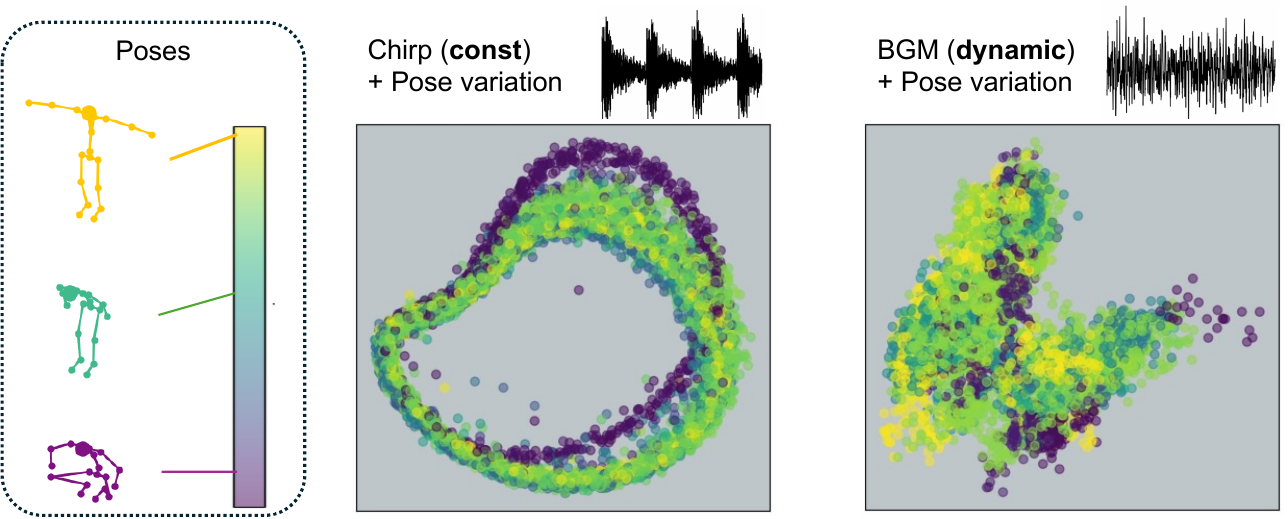}}
\caption{\textbf{Acoustic feature visualization.} (Left) With the repeated chirp signals, differences in collected audio data correspond directly to pose changes, forming patterns for each pose. (Right) In our setting, changes in BGM overshadow human pose variation.
}
\label{fig:differences}
\vspace{-8mm}
\end{figure}

Utilizing everyday BGM as a sensing signal introduces several critical challenges: {\bf(i)} human motion induces slight variations in the phase and amplitude of the collected audio data due to sound attenuation and reflection at the human body surface. In settings where the sensing signal itself varies significantly with music, these pose-related components can be overshadowed by changes in the music. To demonstrate how challenging this task is compared to methods using chirp signals, Fig.~\ref{fig:differences} visualizes the difference between recorded sounds when regular music and repeated chirp signals are used for human pose estimation. In this figure, the common acoustic features (\ie, mel spectrogram and intensity vector) of recorded audio are mapped onto a two-dimensional space using principal component analysis (PCA), with different colors corresponding to different poses. The figure shows that with repeated chirp signals, signal changes are directly associated with changes in poses, forming clear pose patterns even without model training. In contrast, with the regular music we use, it becomes significantly difficult to extract pose information since the changes in poses are largely obscured by the fluctuations in the music. We strongly recommend watching the supplementary material videos with audio to understand the system overview and to grasp the magnitude of variation in the sensing signals.
{\bf(ii)} unlike chirp signals that sweep across a wide range of frequencies, BGM has a limited frequency band that changes over time (see Fig.~\ref{fig:teaser}). Since pose information is observed as changes occurring within the BGM, the model needs to selectively attend to the relevant portions of the spectrogram.
{\bf(iii)} since this framework does not assume specific predefined signals such as chirp, the method needs to adapt to unseen acoustic signals for inference.

To address these challenges, we propose a novel model called \ours.
To overcome the first challenge of separating sound variations caused by the non-periodic source and the human pose, we introduce a Contrastive Pose Extraction module (CPE module). This module effectively promotes the extraction of pose components while simultaneously excluding music components by employing a contrastive loss within a shared feature space. 

To address the second challenge of limited frequency bands and the third challenge of generalizing to unseen musics, we incorporate a Frequency-wise Attention Module (FA module). This module uses an attention mechanism to effectively identify the frequency bands of the spectrogram containing human pose information from the recorded signal, allowing it to extract posture-relevant information even from limited frequency bands or unseen music.

Since no existing work has addressed inferring 3D human poses with BGM, we create Acoustic Music-based Pose Learning (AMPL) dataset, an original dataset for this task. 
Following~\cite{shibata2023listening}, we set up an active acoustic-sensing system using a single pair of ambisonics microphones and loudspeakers. We then actively record the acoustic signals in which BGM and human movement-caused signals are mixed. These signals are synchronized with motion capture~(Mocap) data and captured in an ordinary classroom environment with a certain level of noise.
\revm{
We emphasize that this work presents a first controlled study to establish a baseline for music-based acoustic pose estimation. As an initial step toward practical deployment, the experimental setup is intentionally controlled to enable rigorous quantitative evaluation and fair comparison, while minimizing the influence of environmental factors such as room layout, sensor placement, and background noise.
}

Our contributions are summarized as follows:
(1) For the first time, we propose the 3D human pose estimation model using BGM that blends into daily life;
(2) We introduce a framework that directly maps acoustic signals, in which BGM- and human-movement-caused signals are mixed, to 3D human poses;
(3) We introduce our CPE module, which employs a contrastive loss to encourage the model to extract only pose information from the collected audio, and our FA module, which dynamically calculates attention to focus on specific frequency bands conditioned on sensing sounds; and 
(4) We construct the AMPL dataset, a novel dataset using four BGM tracks spanning two different genres as sensing signals. While preliminary work~\cite{music2pose} explored the feasibility of music-based sensing, it suffered from significant accuracy degradation. We overcome this by introducing the CPE and FA modules, which are essential for isolating pose information from dynamic, non-stationary signals.

\section{Related Work}
\label{sec:Related_work}
Table~\ref{tab:comparison} summarizes where our method is positioned among the existing studies relevant to ours.
This section describes them and highlights their differences with our approach.

\noindent
\textbf{Human Pose Estimation with Different Modalities. }
Estimating human pose is a fundamental task for understanding human behavior. It has attracted substantial interest due to its relevance in real-world applications and ubiquitous computing scenarios.
Many studies have featured RGB-based models, since human pose can be perceived easily through cameras. However, using RGB images or videos can lead to reduced accuracy in cases of occlusion or poor lighting, and the large amount of visual information they contain also raises concerns over privacy~\cite{fredrikson2015model, zheng2023deep}. With the help of advanced deep learning techniques, the use of other modalities, such as WiFi/RF and mmWave, has also been rapidly increasing~\cite{zhao2018through, 
jiang2020towards, ren2022gopose, fan2025diffusion, Li_2022_WACV}. While these modalities can perform well in dark environments, there are some limitations regarding their use in settings with precise instruments, such as in hospitals or during flights. Furthermore, certain materials such as water or metal lead to occlusion issues for these modalities. 
We will address these limitations by utilizing acoustic signals for 3D human pose estimation.

\begin{figure*}[t]
\centering
\centerline{\includegraphics[width=0.9\linewidth]{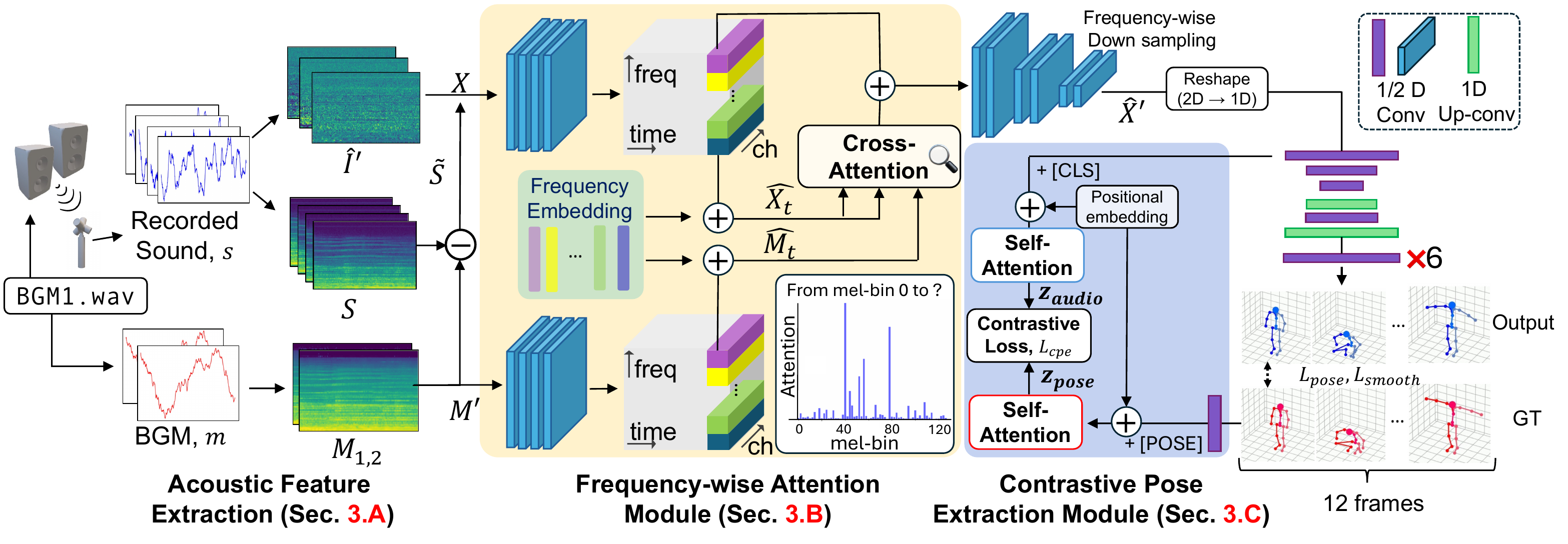}}
\vspace{-2mm}
\caption{The overview of our framework for BGM–based 3D human pose estimation.}
\label{fig:overview}
\vspace{-3mm}
\end{figure*}


\noindent
\textbf{Active Acoustic Sensing. }
Acoustic signals, which have significantly lower frequencies than the signals mentioned above, are subject to fewer usage restrictions. Additionally, acoustic signals, due to their long wavelengths, can transmit information even in environments with obstacles. 
Inspired by bats' spatial recognition abilities, \cite{christensen2020batvision} achieved room-depth estimation based on acoustic signals and two microphones. 
Indoor human position estimation was also conducted using sine sweeps and music~\cite{wang2024soundcam}. Although not focused on sensing, previous studies~\cite{xu2023soundingbodies, huang2024modeling} analyzed the relationship between 3D human poses and the nearby sound field as part of an acoustic scene rendering task.
Regarding the human pose estimation, some prior studies have successfully predicted joint positions using either RGB images and acoustic signals~\cite{posekernellifter} or solely multi-channel acoustic signals~\cite{shibata2023listening,oumi2024acousticbased3dhumanpose}. 
However, all previous works have utilized mathematically optimized chirp signals to investigate echo characteristics, and these systems are impractical considering that chirp signals contain distracting, audible noise within the human hearing range (see Fig.~\ref{fig:teaser} (left)).
Although there has been an attempt to apply music to the existing acoustic-based human pose estimation model, their accuracy significantly degrades compared to when chirp signals are used~\cite{music2pose}.
Other studies tackling active hand/upper-body pose estimation based on acoustic signals required users to put on wearable devices~\cite{amesaka2022gesture, audiotouch, 
handgesture, mahmud2023posesonic}, which imposes extra effort on users and limits the estimation to specific body parts such as the arms. For more practical human sensing, we utilize \revm{non-engineered} BGM for noninvasive active sensing.



\section{Methodology}
\label{methodology}
Our goal is to estimate the 3D human pose sequence $\{\mathbf{p}_t\}_{t=1}^T$ of a target subject standing between a microphone and loudspeakers, given the recorded sound sequence $\{\mathbf{s}_t\}_{t=1}^T$ and original music sequences $\{\mathbf{m}_{i, t}\}_{t=1}^T$ emitted from the $i$-th speakers.
Here, $T$ indicates the length of input and output sequences and $t$ means each timestep. We assume a typical consumer speaker setup with two speakers (\ie, right and left). Following~\cite{shibata2023listening}, we also use one ambisonics microphone, which captures omnidirectional (w) and x, y, and z components with four channels in B-Format. 

An overview of our \ours~method is shown in Fig.~\ref{fig:overview}. 
Our framework includes three main components: an acoustic feature extraction module, the Frequency-wise Attention (FA) module, and the Contrastive Pose Extraction (CPE) module. The FA module enables the model to focus on subtle acoustic variations caused by human posture by dynamically applying attention across frequency bands. The CPE module then uses contrastive learning on the embeddings of recorded audio and pose, which encourages the model to extract pose components from the audio data while eliminating sensing music effects.
Acoustic feature engineering is described in Sec.~\ref{sec:acoustic_feature_extraction}. Secs.~\ref{sec:attention} and ~\ref{sec:cpe} explain our main technical contributions—namely, the FA module and the CPE module, respectively. Then, Sec.~\ref{sec:loss_functions} describes our loss functions.

\subsection{Acoustic Feature Extraction}
\label{sec:acoustic_feature_extraction}
To generate the sequence of the audio feature vectors as the input to our framework, one straightforward way is to directly feed raw signals into some networks for audio feature learning. However, following~\cite{ginosar2019gestures,shibata2023listening}, which showed the efficacy of feature engineering before feeding signals into a network, 
we generate three types of acoustic features. From the recorded sound signals $\mathbf{s}_t$, we extract the intensity vector $\mathbf{I_{\mathrm{intensity}}} \in \mathbb{R}^{3 \times b \times T}$
, including three channels of $(x, y, z)$-directional components, and the log-mel spectrum $\mathbf{S_{\mathrm{logmel}}} \in \mathbb{R}^{4 \times b \times T}$. Here, $b$ denotes the number of frequency bins. These features are often used for sound source localization~\cite{yasuda2020sound} and audio event detection~\cite{kong2019sound}, respectively. Given the original BGM sound signals $\mathbf{m}_{i}$ emitted from the left and right speakers, we generate the log-mel spectrum $\mathbf{M_{\mathrm{logmel}}} \in \mathbb{R}^{2 \times b \times T}$.
Since the range of each signal is different between the intensity vector and log-Mel spectrogram, we standardized them before concatenation.  
In our implementation, we use $b=128$ and $T=12$.

{\bf{The intensity vector}} is the feature that is often used for the direction of arrival (DOA) estimation~\cite{yasuda2020sound, pavlidi20153d} since it represents the acoustical energy direction of signals. It is also used for pose estimation~\cite{shibata2023listening}.

{\bf{The log-mel Spectrum}} is an acoustic time-frequency representation and is known for its good performance as the input of a convolutional neural network~\cite{sakashita2018acoustic, mcdonnell2020acoustic}.
\revm{
While preliminary work~\cite{music2pose} relied on simple log-mel spectrogram subtraction as a preprocessing step without modifying the underlying pose estimation architecture, our framework introduces two dedicated model-level modules to explicitly address the non-stationarity of BGM signals.
}
STFT is performed for the received audio signal $\mathbf{s}_t$ or the original music sound signal $\mathbf{m}_t$, and we apply mel filter bank and logarithmic transformation to compute $\mathbf{S_{k, t}}, \mathbf{M_{k, t}}$.

To filter out the influence of BGM from our recorded audio,
\revm{
 we explicitly subtract the original music data $\mathbf{M}_i$ as the clean reference music signal, emitted from the $i$-th speaker, from the recorded audio data $\mathbf{S}$ after standard normalization. 
Since the sensing system itself controls the playback, it always has direct access to the unmodified playback signal prior to its emission into the room. $\mathbf{M}_i$ is therefore reliably available during both training and inference, and does not require an additional recording device. Sound difference features $\hat{S}$ are calculated as follows: 
}

\vspace{-2mm}
\begin{equation}
\mathbf{\hat{\mathbf{S}}}_{i, c, k, t} = \mathbf{S'}_{c, k, t} - \mathbf{M'}_{i, k,t}.
\end{equation}
\vspace{-3mm}

Here, $c$ denotes the channel, and $S'$ and $M'_{i}$ represent the log-mel spectrograms after channel-wise normalization for recorded sounds and music sounds emitted from $i$-th speaker, respectively. Subtraction in a log scale is equal to division in a linear scale, which means this module calculates the pseudo transfer functions. 
By concatenating the intensity vector and these difference features with two speakers, we obtain the feature $\mathbf{X}$ with $4+4+3=11$ channels in total, which is fed into our network model. 

\subsection{Frequency-wise Attention Module}
\label{sec:attention}
To effectively extract acoustic changes caused by human posture in response to dynamic sensing signals, our method proposes a Frequency-wise Attention Module (FA module). Unlike TSP (chirp) signals designed to emit signals with constant intensity across all frequencies within a specific time frame, the frequency in BGM does not have periodicity and varies over time (see Fig.~\ref{fig:teaser} (left)).  Given that human pose information is observed as changes occurring within the sensing sound, accurately estimating poses requires identifying the frequency bands in the spectrogram that contain the sensing music and are effective for pose estimation. At the same time, the frequency-wise downsampling via convolutional operation at the earlier stages may risk blending information from frequency bands unrelated to the playing sound, potentially leading to reduced accuracy. Therefore, we compute the frequency-wise attention for the input features in accordance with the sensing BGM features while maintaining the original frequency resolution at the earlier stages of the model.

The FA module takes acoustic feature $\mathbf{X}$ and the music feature $\mathbf{M}$ as input. To aggregate local information of the inputs, we independently apply 2D CNNs to both $\mathbf{X}$ and $\mathbf{M}$. Here, we do not apply any downsampling to keep the original frequency resolution. 
Then, we obtain the recorded sound feature $\mathbf{X'}_{t} \in \mathbb{R}^{b \times d}$ and the music feature $\mathbf{M'}_{t} \in \mathbb{R}^{b \times d}$ for each time index $t$, where $d$ is the latent dimension.

Because the attenuation and diffraction characteristics of sound vary depending on frequency, we add learnable shared frequency embedding $\mathbf{F} \in \mathbb{R}^{b \times d}$ to the aforementioned two features, $\hat{\mathbf{X}}_{t} = \mathbf{X'}_{t} + \mathbf{F}$ and $\hat{\mathbf{M}}_{t} = \mathbf{M'}_{t} + \mathbf{F}$.
Then, the attention mechanism is calculated as follows:
\begin{equation}
\text{Attention}(\mathbf{Q}_{t},\mathbf{K}_{t},\mathbf{V}_{t}) = \text{softmax}(\frac{\mathbf{Q}_{t} \cdot \mathbf{K}_{t}^T}{\sqrt{d}})\mathbf{V}_{t},
\end{equation}
where 
$\mathbf{K}_{t} = \hat{\mathbf{M}}_{t}\mathbf{W}^{K}$, 
$\mathbf{Q}_{t} = \hat{\mathbf{X}}_{t}\mathbf{W}^{Q}$, and
$\mathbf{V}_{t} = \hat{\mathbf{X}}_{t}\mathbf{W}^{V}$.
Here, $\mathbf{W}^{K}$, $\mathbf{W}^{Q}$, and $\mathbf{W}^{V} \in \mathbb{R}^{d \times d}$ are the key, query, and value projection matrices, respectively. 
In this frequency-wise attention mechanism, an example of the attention weights from a specific mel-bin to others is visualized in the middle of Fig.~\ref{fig:overview} 
The weights exhibit high sparsity, effectively allowing the model to selectively extract information from frequency bands that are relevant to human pose at each time step.
Since both human movement and music changes are continuous and have little long-term dependency, we do not calculate temporal attention.
To preserve the time consistency brought by upstream 2D CNNs, we leverage a skip connection~\cite{he2016deep} followed by 2D CNNs, which involve the frequency-wise downsampling to get feature $\hat{\mathbf{X'}}$.
Then, $\hat{\mathbf{X'}}$ is fed into a reshaping layer, a time-wise 1D U-Net, and 1D CNNs to output pose $\mathbf{p}$ with a size of $12 \times 63$, which consists of 3D $\times 21$ joints for each of the $12$ frames. For the number of convolutional layers, please refer to Fig.~\ref{fig:overview}.

\begin{figure}[t]
\centering
\centerline{\includegraphics[width=1.0\linewidth]{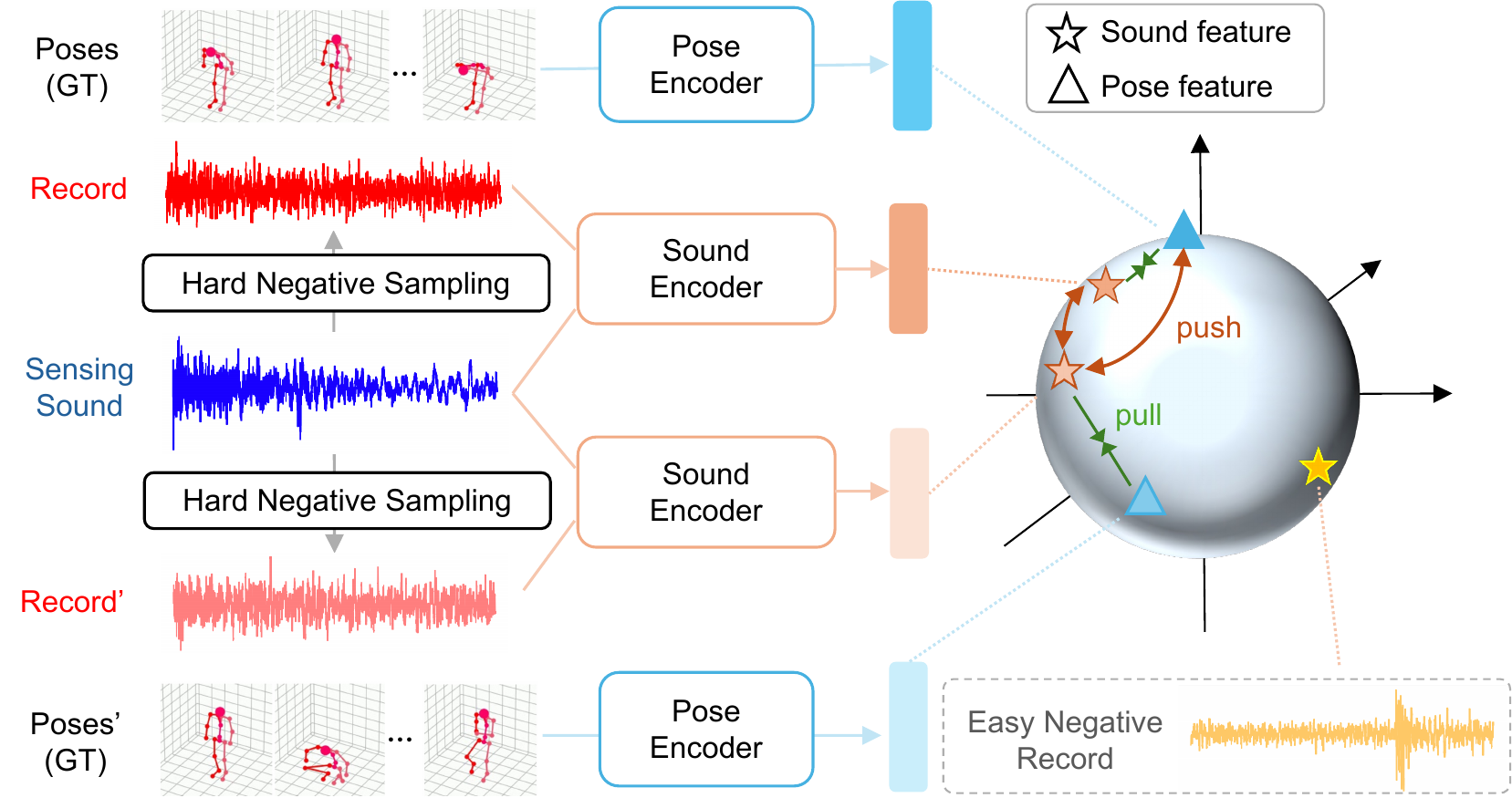}}
\caption{Contrastive learning using human pose and recorded audio is used for BGM-independent feature extraction.}
\label{fig:contrastive}
\vspace{-7mm}
\end{figure}

\subsection{Contrastive Pose Extraction Module}
\label{sec:cpe}

To mitigate the influence of sensing music on the model's representations and to enable the extraction of pose information only, we propose a Contrastive Pose Extraction module (CPE module) based on contrastive learning (Fig.~\ref{fig:contrastive}). This module is designed to incur a high loss when the model’s output is influenced by the sensing BGM. This module performs multi-modal contrastive learning that maps human poses and recorded audio into a common feature space.
In this contrastive learning framework, the model is trained in each minibatch to bring positive pairs (i.e., each pose sequence and its corresponding recorded audio) closer in the feature space while pushing negative pairs farther apart.
When samples recorded with similar sensing music are included in the same mini-batch, the learning process encourages separating them, thereby promoting the extraction of only pose information. Additionally, we propose a novel sampling method called ``BGM-based hard negative sampling." In this approach, similar recordings (and corresponding pose data) obtained using the same sensing BGM are included in the same mini-batch, imposing a more challenging setting for contrastive learning to further enhance the model's discriminative ability (see two similar red and one dissimilar yellow recordings in Fig.~\ref{fig:contrastive}).

For multi-modal contrastive learning, it is necessary to prepare embeddings for both pose sequences and acoustic features. Following the approach of BERT~\cite{devlin2019bert} and Vision Transformer~\cite{dosovitskiy2021imageworth16x16words}, we represent each embedding with a learnable token, [POSE] and [CLS], respectively. For the pose sequence  $\{\mathbf{p}_{t}\}_{t=1}^T$, after adjusting the dimensions through a 1D convolution layer, it is concatenated with the [POSE] token and combined with a learnable time embedding. For acoustic features, as shown in Fig.~\ref{fig:overview}, they are reshaped to 1D and passed through a convolutional layer, then concatenated with the [CLS] token and combined with the time embedding. For each modality, we apply a single layer of self-attention to aggregate temporal information, after which we extract the [POSE] and [CLS] tokens. These outputs are normalized to map onto a hypersphere and expressed as $\mathbf{z_{pose}}$ and $\mathbf{z_{audio}}$, respectively.

With $\langle \cdot, \cdot \rangle$ denoting the inner product and the temperature $\tau$, the cosine similarity $s_{i,j}$ between $i$-th pose and $j$-th audio sequences and our contrastive loss are expressed as follows:
\begin{equation}
s_{i,j} = \langle \mathbf{z_{pose}}_i, \mathbf{z_{audio}}_j \rangle
\end{equation}
\begin{equation}
\scalebox{0.9}{$
L_\mathrm{cpe} = \frac{1}{2N} \sum_{i=1}^{N} \left[ -\log \frac{\exp(s_{i,i}/\tau)}{\sum_{k=1}^{N} \exp(s_{i,k}/\tau)} - \log \frac{\exp(s_{i,i}/\tau)}{\sum_{k=1}^{N} \exp(s_{k,i}/\tau)} \right]
$}
\end{equation}

\subsection{Other Loss Function}
\label{sec:loss_functions}
Similar to previous methods~\cite{jiang2020towards, shibata2023listening}, our network is also trained using loss functions designed to address the correctness and smoothness of the poses. 
The pose loss measures the Mean Squared Error (MSE) between the $j$-th predicted and ground truth joint position of $i$-th subject at timestamp $t$ (\ie, $p_{i, t, j}$ and $\hat{p}_{i, t, j}$, respectively). 
\begin{equation}
\scalebox{1.0}{
$\mathcal{L}_\mathrm{pose}(\theta) = \frac{1}{TNJ}\Sigma^{N}_{i=0}\Sigma^{T}_{t=0}\Sigma^{J}_{j=0}(\hat{p}_{i, t, j} - p_{i, t, j})^{2}$
}
\label{mse_loss}
\end{equation}
Here, $\mathcal{L}$ denotes the loss function, $\theta$ represents trainable parameters and $J$ is the total joint size.
We also use the smooth loss to make our prediction smoother.
\begin{equation}
\scalebox{0.95}{
$\mathcal{L}_\mathrm{smooth}(\theta) = \frac{1}{(T-1)}\Sigma^{T}_{t=2}\text{MSE}((\hat{\mathbf{p}}_{t} - \hat{\mathbf{p}}_{t-1}) - (\mathbf{p}_{t} - \mathbf{p}_{t-1})).$
}
\label{smooth_loss}
\end{equation}
With weight parameters $w_{\alpha}$ and $w_{\beta}$, our total loss function is as follows:
\begin{equation}
L =  L_\mathrm{pose} + w_{\alpha} L_\mathrm{smooth} + w_{\beta} L_\mathrm{cpe}
\label{totalloss}
\end{equation}

\begin{table}[tb]
\centering
\caption{Details of the proposed AMPL dataset.}
\scalebox{0.8}{
\begin{tabular}[h]{lccccc}
\toprule
Name & Genre & Duration (min:sec) & Subject IDs\\ 
\midrule 
ARNOR~\cite{arnor} & Ambient & 11:12 & [1, \red{1}, 2, 3, 4, 5, \red{9}]\\
Cirrus~\cite{cirrus} & Ambient & 6:58 & [1, \red{1}, 2, 3, 4, 5, 6, 7, 8, \red{9}]\\
MANTRA~\cite{mantra} & Ambient & 8:37 & [1, \red{1}, 2, 3, 4, 5, \red{9}]\\
Kurina blues~\cite{jazz} & Jazz & 14:44 & [6, 7, 8]\\
\bottomrule
\label{tab:data_stats}
\vspace{-5mm}
\end{tabular}
}
\end{table}

\begin{figure}[t]
\centering
\vspace{-3mm}
\centerline{\includegraphics[width=0.8\linewidth]{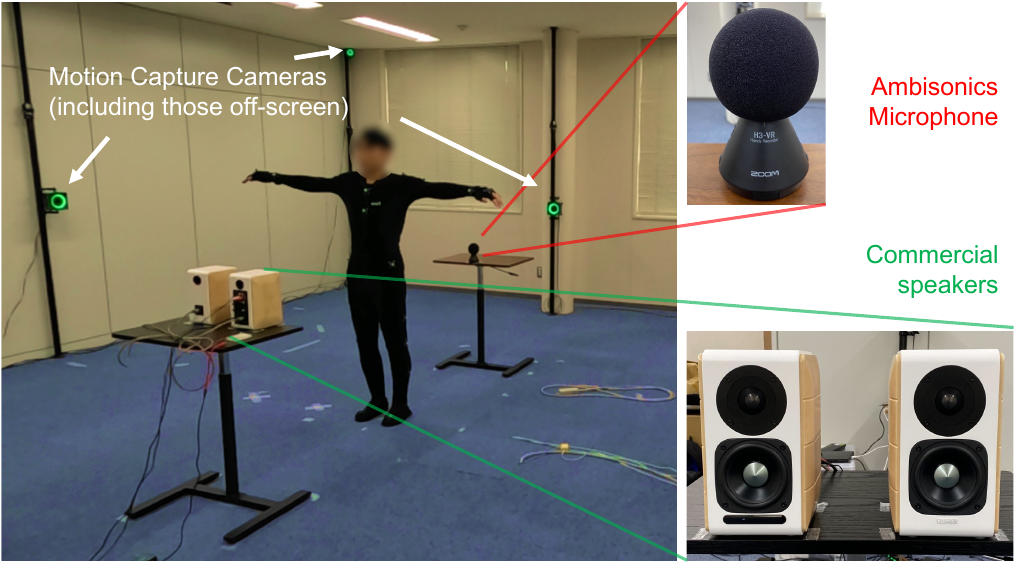}}
\caption{The setup of our experiments. 
}
\label{fig:environments}
\vspace{-5mm}
\end{figure}

\section{Human Pose Dataset with Music}
For our task, we created a novel dataset, \textbf{AMPL} (\textbf{A}coustic \textbf{M}usic-based \textbf{P}ose \textbf{L}earning dataset), which links musical sensing signals with 3D human poses. As shown in Table~\ref{tab:data_stats}, our AMPL dataset contains over 4 hours of synchronized acoustic and motion data, consisting of 27 recording sessions from 9 subjects across four different BGM tracks. These data are captured in a classroom environment, which includes background noise and reverberation, using motion capture (Mocap) and 16 cameras. For the right and left speakers, we used Edifier ED-S880DB, commercial high-resolution speakers. Additionally, we captured acoustic signals using an ambisonics microphone (Zoom H3-VR). As shown in Fig.~\ref{fig:environments}, the speakers and microphone were placed one meter away from the subjects, who were positioned in the line of sight. All recordings were conducted under a protocol that obtained written informed consent from \revm{all participants}. 


Table~\ref{tab:data_stats} summarizes detailed information on the BGM used for sensing, including titles, genres, duration, and target subjects. 
The subjects performed various poses, such as standing still, bending, and walking. We shifted and normalized each joint position so that the hips were positioned at the origin of our coordinates, and the spine-hip distance was set as 1. To investigate the effect of the sound absorption properties of clothing, two subjects wore their personal outfits (the subject IDs indicated in \red{red}). 

\section{Experimental Settings}
\label{experimental_settings}
\noindent
\textbf{Baselines. }
So far, no existing method has performed 3D human pose estimation using both music and recorded data as inputs.
Therefore, we compared our model against the following approaches that are similar to ours;
(i) \textbf{Jiang \etal~\cite{jiang2020towards}} that uses WiFi signals, but like us, they use actively acquired low-dimensional data as input.
(ii) \textbf{Ginosar \etal~\cite{ginosar2019gestures}}, which uses acoustic modality, similar to our approach, but focuses on gesture generation with audio that contains semantics, such as human speech. 
(iii) \textbf{Shibata \etal~\cite{shibata2023listening} and Nakamura \etal ~(its music-based extension)~\cite{music2pose}}. While \cite{music2pose} applies subtraction-based preprocessing to the model in ~\cite{shibata2023listening} to handle music, we propose specialized architectural enhancements: the CPE and FA modules. These modules explicitly address the non-stationarity of BGM at the model level, rather than through preprocessing alone. We use the architecture from ~\cite{shibata2023listening} as our primary baseline to highlight the effectiveness of these structural improvements.
\textbf{Please note that all baseline models were trained on our dataset for fair architecture comparisons}. To this end, we modified the input and final layers of each baseline network to accept our acoustic signals as input and output 3D human poses.

\vskip\baselineskip
\noindent
\textbf{Evaluation Metrics. }
Our quantitative metrics are as follows: root mean square error (RMSE), mean absolute error (MAE), and percentage of correct key point (PCK). RMSE and MAE are used to express the average displacement between the predicted and the true joint positions, while PCK indicates the percentage of the model's predictions that fall within a predefined threshold from the correct positions. 
Regarding PCKh, we use the PCKh@0.5 score with a threshold of 50\% for the head–neck bone link.

\vskip\baselineskip
\noindent
\textbf{Implementation Details. }
To synchronize the playback music with the collected sound data, the playback music were upsampled from 44,100 Hz to 48,000 Hz. Regarding the acoustic features, sampling was conducted to achieve a frame rate of 20 FPS. 
For training, we used Adam~\cite{adam} for optimization and employed a cosine annealing learning scheduler~\cite{cosine_annealing}, which reduced the learning rate from 0.003 to 0.001. All training was conducted using exponentially moving average for model updates. The batch size was set to 64. All models were trained for 30 epochs. Regarding the weights in Eq.~\ref{totalloss}, we set $w_{\alpha}=100$ and $w_{\beta}=1$. In our main experiments, the temperature $\tau$ is set to 0.07. 
In our model, all prediction results were computed using sliding window inference on sequences consisting of 12 frames. Specifically, a stride of 4 frames was set, and overlapping regions were weighted using a Gaussian distribution centered at half the sequence length with its standard deviation set 6. For a fair comparison, this post-process was applied to all methods, including the existing baselines.

\begin{table*}[tb]
\footnotesize
    \centering
      \caption{Results in the (a) single-music and the (b) cross-music settings \revm{with standard deviation}. All results are obtained in a cross-subject setting.}
      \vspace{-7mm}
        \begin{center} 
        \resizebox{0.75\linewidth}{!}{ 
        \begin{tabular}[t]{@{\hskip 1mm}lcccc@{\hskip -1mm}ccc@{\hskip 1mm}ccc@{\hskip 1mm}ccc@{\hskip 1mm}ccc}
        \toprule
        & \multicolumn{3}{c}{(a) Single-Music} & & \multicolumn{3}{c}{(b) Cross-Music}\\
        \cmidrule{2-4} \cmidrule{6-8}
        {Method} & 
        RMSE ($\downarrow$) & MAE ($\downarrow$) & PCKh@0.5  ($\uparrow$) & & 
        RMSE ($\downarrow$) & MAE ($\downarrow$) & PCKh@0.5  ($\uparrow$) 
        \\
        \midrule
        Jiang \etal~\cite{jiang2020towards} & 1.338 \revm{(0.042)} & 0.768 \revm{(0.033)}& 0.251 \revm{(0.034)}& & 1.417 \revm{(0.025)}& 0.800 \revm{(0.012)}& 0.272 \revm{(0.008)}\\
        Ginosar \etal~\cite{ginosar2019gestures} & 1.223 \revm{(0.026)}  & 0.666 \revm{(0.016)}& 0.379 \revm{(0.023)}& & 1.274 \revm{(0.022)}& 0.682 \revm{(0.011)}& 0.375 \revm{(0.017)}\\
        Shibata \etal~\cite{shibata2023listening} &1.090 \revm{(0.024)}& 0.574	\revm{(0.016)}& 0.468 \revm{(0.007)}& &  1.110 \revm{(0.024)}& 0.556 \revm{(0.014)}&	0.499 \revm{(0.008)}\\
        \rowcolor{LightRed} Ours & \textbf{0.923 \revm{(0.013)}} & \textbf{0.453 \revm{(0.006)}} & \textbf{0.573 \revm{(0.005)}} & & \textbf{1.036 \revm{(0.033)}} & \textbf{0.494 \revm{(0.018)}} & \textbf{0.570 \revm{(0.011)}}\\
        \bottomrule
        \label{tab:quantitative}
        \end{tabular}
        } 
        \end{center} 
\vspace{-9mm}
\end{table*}

\begin{table}[tb]
\centering
\caption{Cross-genre setting results \revm{with standard deviation} (train=ambient, test=jazz).}
\scalebox{0.9}{
\begin{tabular}[h]{lccccc}
\toprule
Method & RMSE ($\downarrow$) & MAE ($\downarrow$) & PCKh@0.5 ($\uparrow$)\\ 
\midrule 
Jiang \etal~\cite{jiang2020towards} & 1.418 \revm{(0.080)}& 0.849 \revm{(0.080)}& 0.221 \revm{(0.073)}\\
Ginosar \etal~\cite{ginosar2019gestures} & 1.326 \revm{(0.073)}& 0.727 \revm{(0.041)}& 0.360 \revm{(0.002)}\\
Shibata \etal~\cite{shibata2023listening} & 1.112 \revm{(0.020)}& 0.595 \revm{(0.015)}& 0.376 \revm{(0.019)}\\
\rowcolor{LightRed}
Ours & \textbf{1.065 \revm{(0.070)}} & \textbf{0.543 \revm{(0.041)}} & \textbf{0.463 \revm{(0.026)}} \\
\bottomrule
\label{tab:cross_genre}
\vspace{-7mm}
\end{tabular}
}
\end{table}

\section{Experimental Results}
\label{experimental_result}
We conducted four complementary experiments to evaluate the proposed method from algorithmic, practical, and interpretability perspectives:
    (A) \textbf{Baseline comparison:}
    We benchmarked our model against the aforementioned baselines to establish overall performance gains; 
    (B) \textbf{Ablation study:}
    We removed the CPE and FA modules in turn to quantify the individual contribution of our two main technical contributions;
    (C) \textbf{Plain-clothes generalization:}
    Although the model is trained on motion-capture-suit data with ground-truth poses, we evaluate its accuracy on the \emph{In Plain Clothes} dataset to confirm applicability when subjects wear everyday attire; and
    (D) \textbf{Noise robustness:}
    To assess performance in realistic conditions, we test the model 
    after adding synthetic noise to the input audio.

\begin{figure}[t]
\centering
\centerline{\includegraphics[width=0.9\linewidth]{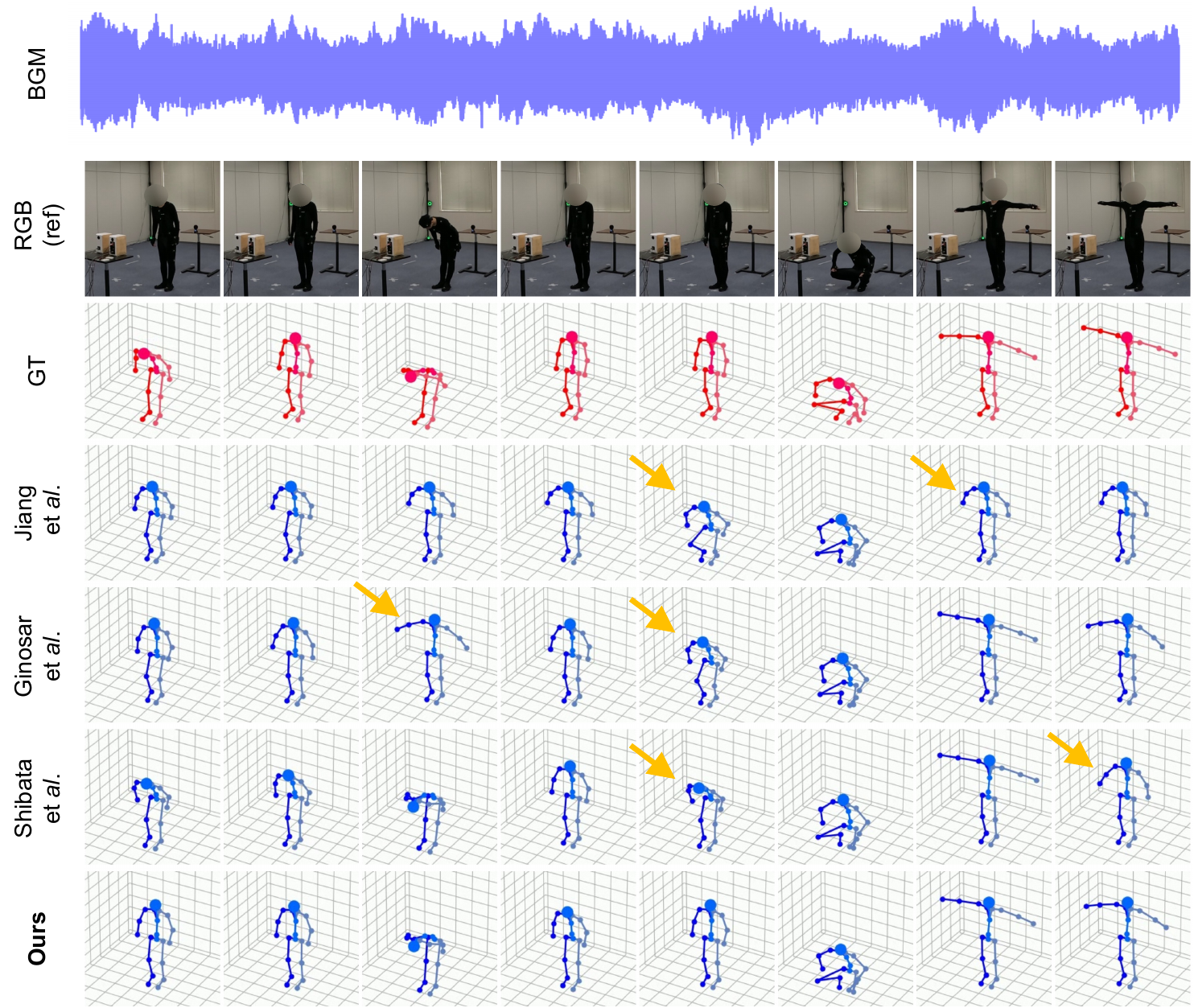}}
\caption{The qualitative results with the Mocap suit under the single-music setting.}
\label{fig:mocap_qualitative}
\vspace{-4mm}
\end{figure}
\begin{table}[t]
    \centering
      \caption{Ablation study in the single- and the cross-music settings.}
      \vspace{-3mm}
        \scalebox{0.85}{
        \begin{tabular}[t]{@{\hskip 1mm}lcccc@{\hskip -1mm}ccc@{\hskip 1mm}ccc@{\hskip 1mm}ccc@{\hskip 1mm}ccc}
        \toprule
        & \multicolumn{3}{c}{(a) Single-Music} & & \multicolumn{3}{c}{(b) Cross-Music}\\
        \cmidrule{2-4} \cmidrule{6-8}
        \multirow{3}{*}{Method} & 
        \multirow{2}{*}{RMSE} & \multirow{2}{*}{MAE} & \multirow{2}{*}{\shortstack[c]{PCKh \\ @0.5}} & & 
        \multirow{2}{*}{RMSE} & \multirow{2}{*}{MAE} & \multirow{2}{*}{\shortstack[c]{PCKh \\ @0.5}} \\
        \\
        & ($\downarrow$) & ($\downarrow$) & ($\uparrow$) & & ($\downarrow$) & ($\downarrow$) & ($\uparrow$)\\ 
        \midrule
        \rowcolor{LightRed}
        Ours & \textbf{0.923} & \textbf{0.453} & \textbf{0.573} & & 1.036 & \textbf{0.494} & \textbf{0.570}\\
        w/o CPE module & 0.945 & 0.481 & 0.531 & & \textbf{1.024} &	0.501 &	0.547 \\
        w/o BGM Hard Negative & 0.928 & 0.457 & 0.566 & &1.065 & 0.506 & 0.559 \\
        w/o FA module & 1.025 & 0.509 &	0.536 & & 1.210 & 0.593 &0.512\\
        w/o BGM conditioning & 0.970 & 0.484 &0.537 & & 1.075 & 0.522 & 0.543\\
        \bottomrule
        \label{tab:ablation}
        \vspace{-8mm}
        \end{tabular}
        }
\end{table}

\subsection{Comparison with Other Baselines}
\label{quantitative_results}
We evaluated our proposed method against the aforementioned models within the following three distinct scenarios: (a) a single-music setting, wherein the same ambient BGM was used as the acoustic source for both training and testing; (b) a cross-music setting, in which two  ambient and one jazz BGM were used for training and the remaining ambient music was used for testing only; and (c) a cross-genre setting, in which all ambient tracks were used for training and one jazz BGM was used for testing. Regarding (a) and (b), only ambient tracks were used for testing since they have few silence intervals and are suitable for stable evaluation compared to jazz BGM. 
(b) and (c) are settings designed to evaluate the model's generalization performance to unseen music and to both unseen music and genres, respectively.


The relationship between the music used for data collection and the subjects is shown in Table~\ref{tab:data_stats}. For experimental settings (a) and (b), we conducted accuracy evaluations using K-fold cross-validation (K=5) on subjects with IDs 1–5, who completed data collection with all prepared ambient audio sources. The remaining three subjects, who collected data with ``Cirrus'' and ``Jazz'', were also included in the training data for the Single-Music setting with ``Cirrus'' and for the Cross-Music setting with all three ambient tracks. In the Cross-Genre setting, we conducted three-fold cross-validation, using subjects [6, 7, 8] for evaluation and the remaining subjects' data for training. Please note that across all BGM settings, we evaluated the model's accuracy on a \textbf{cross-subject} setting. For evaluation settings (a) and (b), the reported accuracy is the average across the three ambient music tracks. Additionally, to facilitate more effective accuracy comparisons, all experiments report accuracies averaged over three random seeds. 

Table~\ref{tab:quantitative} summarizes the quantitative results in both the single- and cross-music settings. We can see that our proposed method outperformed the previous models in all metrics under both settings, highlighting that it has a strong capacity to capture 3D human poses based on music sounds. Please note that the values in the table are normalized, with the hip-spine distance set to 1. In our Mocap system, with the spine position closest to the hip, the average of this distance across the 8 subjects was 9.21 cm. In particular, regarding PCKh@0.5, the proposed method outperforms the second-best approach by approximately 10\% and 7\% in both settings, suggesting that our method accurately identifies challenging poses, such as the ``T-pose''.


Table~\ref{tab:cross_genre} shows the results when three ambient music tracks are used for training while jazz music is used for testing. Jazz exhibits more drastic changes in pitch and volume compared to ambient music, and it includes silent intervals during which the sensing signal does not work. Consequently, we observe a decrease in accuracy in PCKh@0.5 compared to evaluations conducted with ambient music (Table~\ref{tab:quantitative}). However, even so, our method records significantly higher accuracy than existing methods, suggesting that our approach is robust across different music genres.


Fig.~\ref{fig:mocap_qualitative} shows the qualitative results for every 10 seconds under the single-music setting. 
The baseline methods failed to predict various poses, as shown by the yellow arrows, partly due to the effects of dynamic BGM. 
On the other hand, our method successfully predicted stable pose sequences, including T-pose, which had a very small \revm{reflective} area.

\begin{figure}[t]
\begin{center}
\vspace{-3mm}
\includegraphics[width=\linewidth]
{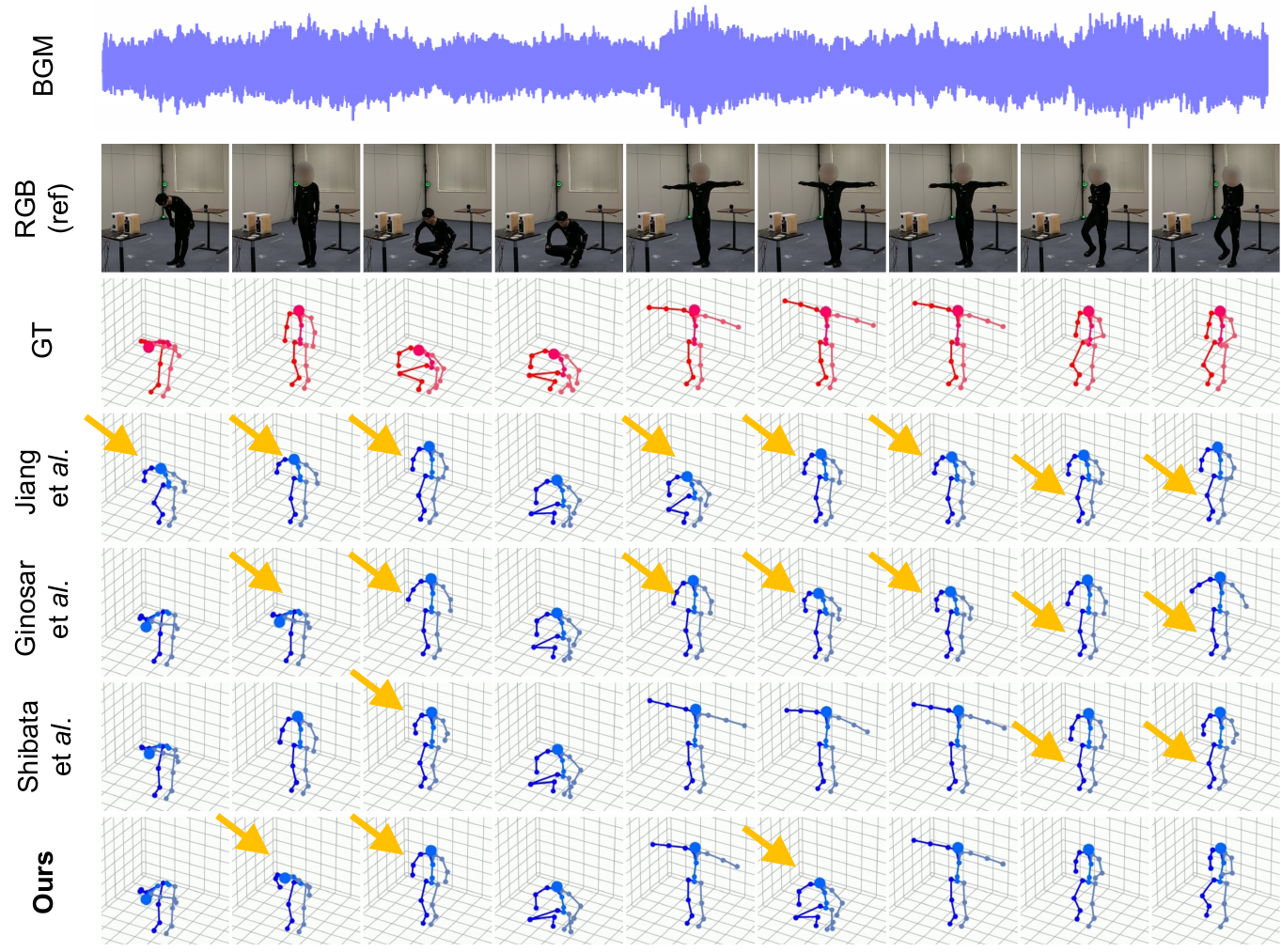}
\end{center}
\vspace{-2mm}
\caption{\revm{The qualitative results in the cross-music setting. }}
\label{fig:qualitative_cross}
\vspace{-4mm}
\end{figure}

\revm{
Fig.~\ref{fig:qualitative_cross} shows the qualitative results in the cross-music setting.
We can see that our model successfully reduced false predictions compared with other baselines,
showcasing the proposed model is robust for BGMs which are not included in training data. Notably, the results from the two rightmost columns show that only the proposed method successfully recognizes walking, which requires high spatial resolution. 
}

\subsection{Ablative Analysis}
We investigated the effects of our main technical contributions, \ie, the CPE module and the FA module. 
For the CPE module, we also prepared a setting in which mini-batches are created randomly without using proposed BGM-based hard negative sampling. Furthermore, to demonstrate the effectiveness of using playback music as additional input in our model, we conducted accuracy evaluations in a setting where we removed the playback BGM channels (right/left) from the inputs and calculated FA module based on self-attention.
Table~\ref{tab:ablation} shows that both our proposed modules contributed to improving the estimation accuracy. 
\revm{
While both modules are beneficial, the FA module has a substantially larger quantitative impact. Removing the FA module increases RMSE from 0.923 to 1.025 in the single-music setting and from 1.036 to 1.210 in the cross-music setting, whereas removing the CPE module results in a smaller degradation. This suggests that adaptive frequency selection is the dominant factor for handling non-stationary 
BGM, while the CPE module provides complementary gains.
}



\subsection{Evaluation with the In Plain Clothes Dataset}
\begin{figure}[tb]
\centering
\centerline{\includegraphics[width=1.0\linewidth]{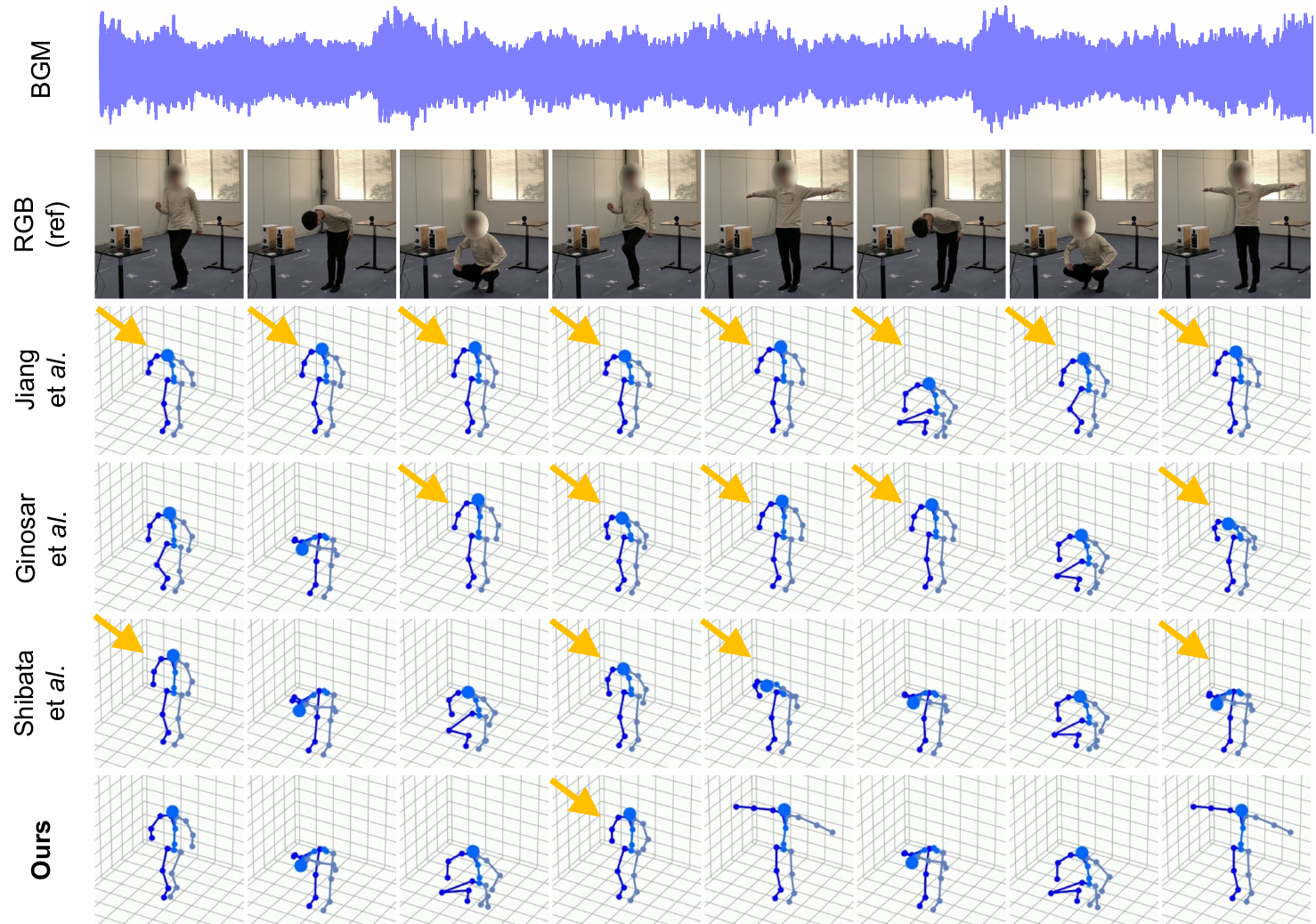}}
\caption{The qualitative results in the cross-music setting with the subjects wearing plain clothes.}
\label{fig:wo_mocap_qualitative}
\end{figure}

Fig.~\ref{fig:wo_mocap_qualitative} shows the qualitative results for every 16 seconds under the cross-music setting with the subject wearing plain clothes. As highlighted by the yellow arrows, we can see that our proposed method significantly reduced false predictions, compared to the baselines. The results indicate that the proposed method was robust to variations in the clothing worn by the subjects, although all the training data were captured with subjects putting on Mocap suits.

\subsection{Noise Robustness Investigation}
It is important to investigate the noise robustness of the method when considering applications in real-world environments.
Table~\ref{tab:snr_atten} shows the results in a single-music setting with Gaussian noise added to the collected sound data of both the training and test datasets. Here, SNR (Signal-to-noise ratio) is 10. Our method showed robustness to noise, achieving significantly higher estimation accuracy than existing methods by a large margin.

\revm{
To further investigate practical noise robustness, we added audio from the DESED dataset~\cite{turpault2019sound} (natural indoor sounds like speech and alarms) into our recorded data as noise. We tested the model’s robustness under two conditions:
(i) Only the test data were mixed with DESED audio,
(ii) Both training and testing data contain randomly sampled DESED audio. 
From the accuracy comparison between Table~\ref{tab:desed} and Table~\ref{tab:quantitative} (a:single-music setting), we see that our model is robust to environmental noise.
In the most challenging zero-shot scenario (i), our model matched the PCKh scores of some noiseless baselines in Table~\ref{tab:quantitative}(a). Also, Table~\ref{tab:desed}(ii) shows that training with noise significantly boosts test-time noise resilience.
Because human voices and environmental sounds differ acoustically from music, our model, which detects changes in the sensing music, can still perform robust inference in such noisy environments.
}

\begin{table}[tb]
\centering
\caption{Noisy environment setting with Gaussian noise (SNR=10).}
\scalebox{1.0}{
\begin{tabular}[h]{lccccc}
\toprule
Model & RMSE($\downarrow$) & MAE($\downarrow$) & PCKh@0.5(\revm{$\uparrow$})\\ 
\midrule 
Jiang \etal~\cite{jiang2020towards} & 1.567 & 0.931 & 0.095\\
Ginosar \etal~\cite{ginosar2019gestures} &1.389 & 0.796 & 0.262\\
Shibata \etal~\cite{shibata2023listening} & 1.328 & 0.737 & 0.354 \\
Ours & \bf{1.202} & 	\bf{0.637} & \bf{0.430}\\
\bottomrule
\label{tab:snr_atten}
\vspace{-7mm}
\end{tabular}
}
\end{table}

\begin{table}[t]
\centering
\caption{\revm{Robustness against natural noises.}}
\vspace{-5mm}
\scalebox{0.9}{
    \begin{tabular}[t]{@{\hskip 1mm}ccc@{\hskip -1mm}ccc@{\hskip 1mm}ccc@{\hskip 1mm}ccc@{\hskip 1mm}ccc}
    \toprule
    \multicolumn{3}{c}{\revm{(i) Zero-shot Testing}} & & \multicolumn{3}{c}{\revm{(ii) Training and Testing}}\\
    \cmidrule{1-3} \cmidrule{5-7}
    \revm{RMSE} & \revm{MAE} & \revm{PCKh@0.5} & & 
    \revm{RMSE} & \revm{MAE} & \revm{PCKh@0.5} \\
    \midrule
    \revm{1.60} & \revm{0.842} & \revm{0.377} & & \revm{1.13} & \revm{0.567} & \revm{0.502}\\
    \bottomrule
    \label{tab:desed}
    \vspace{-7mm}
    \end{tabular}
} 
\end{table}

\section{Discussion and Limitations}

\revm{

\noindent
\textbf{Sensitivity to environmental noise.} As shown in Table 6, performance decreases in the presence of additive noise. Although the proposed method is more robust than the baseline methods and maintains a clear performance advantage, environmental noise remains a challenge for deployment in highly noisy real-world environments.

\noindent
\textbf{Left-right ambiguity during walking.}
The current system sometimes confuses the left and right sides during walking actions. Improving the discrimination of laterally symmetric body motions remains an important direction for future work.

\noindent
\textbf{Line-of-sight requirement.} The current system requires the subject to be positioned in the line of sight between the speakers and the microphone, which limits deployment to scenarios where such a physical configuration is feasible (e.g., dedicated fitness coaching rooms, meeting rooms). This constraint arises because we established a controlled baseline to maximize acoustic reflectivity from the human body. Prior work using chirp signals has demonstrated pose estimation over wider areas including non-LOS conditions through adversarial learning~\cite{oumi2024acousticbased3dhumanpose}, and extending our method in this direction is a promising avenue for future work.

\noindent
\textbf{Dependence on spectral density of sensing music.} Our FA module relies on the sensing music to provide active frequency bands from which pose-related acoustic variations can be extracted. Accordingly, the proposed system may not function reliably with acoustically sparse or narrow-band music (e.g., a slow solo piano piece with extended silences or a very limited frequency range). In such cases, there are insufficient active frequencies for the model to extract pose information. Our experiments used ambient and jazz music, which provide relatively rich spectral content; extending the method to handle a broader range of spectral characteristics is an important direction for future work.

\noindent
\textbf{Dataset diversity.} Although the AMPL dataset constitutes an important first step toward benchmarking BGM-based 3D human pose estimation and provides over four hours of synchronized data across 9 subjects and 4 music tracks in a classroom environment, it remains limited in terms of subject diversity, room acoustic variety, and musical genres. Stronger generalization would require evaluation across multiple environments, a larger and more diverse subject pool, and a broader variety of music characteristics.

\noindent
\textbf{Dependence on a known clean playback signal. }
The current framework requires access to the clean playback-music signal synchronized with the recorded audio. This requirement is acceptable in many scenarios because the system itself plays the music through speakers. Handling unknown or uncontrolled  music without a clean reference signal remains an important direction for future work.
}
\section{Conclusion}
This paper proposes \ours, a human 3D pose estimation that uses BGM as active sensing signals. Unlike existing methods that utilize specific chirp signals which is uncomfortable for humans, our approach uses \revm{pre-selected} everyday music, offering greater flexibility and practicality.
This task is inherently challenging because the amplitude and pitch of the signals vary over time, and the human pose clues in recorded sounds are concealed by music changes. 
Moreover, the effective frequency range of BGM is limited.  
Therefore, we proposed a novel model incorporating the Frequency-wise Attention Module and the Contrastive Pose Extraction Module to focus on the changes in the measurement sounds caused by human pose variations at each moment. 
The proposed method demonstrated the ability to accurately estimate human 3D poses across a wide range of conditions, including unseen test music not used in the training data and unseen subjects wearing plain clothes.
We hope that our research will lead to the development of human sensing applications using practical sounds.

{
    
    \bibliographystyle{IEEEtran}
    \bibliography{ref}
}

\vfill\pagebreak

\end{document}